# Discovery Learning accelerates battery design evaluation

Jiawei Zhang[1,2], Yifei Zhang[2], Baozhao Yi[2], Yao Ren[3], Qi Jiao[3], Hanyu Bai[1,2], Weiran Jiang[3]* and Ziyou Song[1,2]*

[1]*Department of Electrical Engineering and Computer Science, University of Michigan, Ann Arbor, MI, USA.*

[2]*Department of Mechanical Engineering, National University of Singapore, Singapore, Singapore.*

[3]*Farasis Energy USA, Inc., Hayward, CA, USA.*

*Corresponding authors. Email: wjiang@farasis.com, ziyou@umich.edu

**Abstract**

Fast and reliable validation of novel designs in complex physical systems such as batteries is critical to accelerating technological innovation. However, battery research and development remain bottlenecked by the prohibitively high time and energy costs required to evaluate numerous new design candidates, particularly in battery prototyping and life testing[1,2]. Despite recent progress in data-driven battery lifetime prediction, existing methods require labeled data of target designs to improve accuracy and cannot make reliable predictions until after prototyping, thus falling far short of the efficiency needed to enable rapid feedback for battery design[3,4]. Here, we introduce Discovery Learning (DL), a scientific machine-learning paradigm that integrates active learning[5], physics-guided learning[6], and zero-shot learning[7] into a human-like reasoning loop, drawing inspiration from learning theories in educational psychology. DL can effectively learn from historical battery designs and actively reduce the need for prototyping, thus enabling rapid lifetime evaluation for previously unobserved material-design combinations without requiring additional data labeling. To test DL, we present industrial-grade battery data comprising 123 large-format (73–84 Ah) lithium-ion pouch cells based on five years of development and validation in the battery manufacturing industry. This test dataset spans eight material-design combinations—including ultrahigh-nickel cathodes—and diverse cycling protocols, such as realistic electric-vehicle driving profiles. Trained solely on public datasets of small-capacity (1.1–3.5 Ah) cylindrical cells, which differ significantly from our data, DL achieves 7.2% test error in predicting the average cycle life under unknown device variability, using only the first 50 cycles from just 51% of cell prototypes. Under conservative assumptions, this results in savings of 98% in evaluation time and 95% in energy consumption compared to conventional industrial practices. This work highlights the potential of uncovering insights from historical designs to inform and accelerate the development of next-generation battery technologies. DL represents a key advance toward efficient data-driven modeling and helps realize the promise of machine learning for accelerating scientific discovery and engineering innovation[8].

**Introduction**

The world is far off track to meet the United Nations Sustainable Development Goals by 2030[9], such as developing sustainable energy[10]. Accelerating progress toward these goals requires doubled efforts from scientists and engineers[11,12]. Yet, the pace of disruptive innovations in science and technology has slowed over the years[13]. For example, progress in battery technology has happened incrementally with few major breakthroughs since the invention of lithium-ion batteries[14,15]. Looking back at the historical development of scientific and engineering disciplines reveals that time-consuming and resource-intensive experimentation for validating new hypotheses or technologies has caused bottlenecks in a wide range of fields, such as material discovery[16], drug development[17], biology[18], energy[4], and transport[19]. Furthermore, the vastness of the design and sample space poses a significant challenge for scientific validation. Therefore, purely experiment-driven scientific discovery workflows can no longer keep pace with the growing demand for accelerated progress. A paradigm shift in how we evaluate new scientific hypotheses and technologies is urgently needed.

Developing long-life[20–22] batteries is essential to meeting the increasing demand for electric vehicles (EVs) and grid storage. Fast and reliable lifetime evaluation of a large number of new battery designs is crucial for accelerating the development of next-generation batteries. Unfortunately, conducting even a single degradation experiment for battery life testing can take years of painstaking effort[1,23]. Furthermore, battery prototyping and life testing require large amounts of energy even when material mining and refining are excluded[2], resulting in substantial greenhouse gas emissions. Despite advances in manufacturing technologies[2,24] and high-throughput testing[25–27], battery prototyping and life testing remain necessary links in industrial battery development, leaving the prohibitively high time and energy costs unresolved. Failure to address these challenges would seriously hinder the pace of battery innovation and create an intractable 'sustainability dilemma'—advancing battery technologies for sustainable energy could become unsustainable.

Accurate and efficient battery lifetime prediction would offer a shortcut for lifetime evaluation in battery design. The development of forecasting approaches has proceeded along two routes, driven by either physics-based models or by data. Physics-based approaches integrate governing equations with complex degradation models. Although theoretically appealing, the incomplete and still-debated understanding[28] of degradation mechanisms remains a fundamental gap. Data-driven approaches[3,29–32] have provided a mechanism-agnostic alternative in recent years, thanks to the rapid development of artificial intelligence (AI) and machine learning (ML)[33]. However, to improve forecast accuracy of new battery designs, conventional data-driven prediction requires additional degradation experiments for training, and does not effectively leverage historical battery designs. Moreover, conventional data-driven approaches cannot make accurate predictions for target battery designs until after prototyping, causing critical efficiency bottlenecks—particularly when scaling to large design spaces. Owing to the prohibitive upfront training costs[4] and immediate inference costs in terms of evaluation time and energy consumption[1,2,23,34], existing approaches fall far short of the efficiency required to support fast-evolving battery innovation.

Beyond the battery field, using ML to predict battery lifetime is a representative task in the emerging field of 'AI for Science' (AI4Science)[8]; AI and ML can enable reliable scientific validation by offering accurate and efficient predictions. Such scientific prediction problems suffer from challenges in obtaining reliably labeled datasets for ML models, which can involve time-consuming and resource-intensive experimentation. We refer to this challenge as 'data unaffordability'. The Battery Data Genome[35] initiative was proposed in 2022 to address this issue, aiming to unite the global battery community in building extensive and robust datasets. Yet, this moonshot plan will require time to deliver results due to the need for long-term collaboration among battery researchers, cell manufacturers, and policymakers. In addition to data unaffordability, ML-enabled scientific prediction is also hindered by the challenge of 'distribution shift'. By 'distribution shift', we mean that the distribution of training data differs from that of test data, which can negatively impact the generalization performance of ML models. This is a prevalent issue in many disciplines, as scientific discovery inherently involves out-of-distribution efforts[36].

A comprehensive literature search in AI4Science reveals three main strategies for addressing these challenges. First are physics-aided approaches that integrate domain knowledge with ML models[6,37], such as physics-constrained loss functions[38,39] and knowledge-integrated model architectures[18,40]. However, these methods are not well-suited for battery lifetime prediction, as they either require micro-scale physical information, which is not easily accessible, or are specific to certain scientific domains. The second approach uses scaled-up large datasets to enhance the predictive capability of ML models, benefiting from long-term accumulation[36,41–44] or efficient collection[45–49]. Unfortunately, this is not practical for battery lifetime prediction due to data unaffordability. The third approach is active learning[5], which aims to use as few labeled samples as possible by querying informative unlabeled samples for labeling through experimentation[16,50,51]. However, obtaining even a single battery life label remains time-consuming. In summary, it remains difficult for existing approaches in AI4Science to address the dual challenges of data unaffordability and distribution shift in battery lifetime prediction.

Here we propose Discovery Learning (DL), a new ML concept and paradigm for scientific prediction, inspired by human learning theories developed 65 years ago in educational psychology[52]. We show that DL can accurately and efficiently predict the lifetime performance of previously unobserved battery designs under unknown device variability. We demonstrate the effectiveness of DL using both well-accepted public datasets and our large-format battery aging data that have been validated in the battery manufacturing industry instead of lab-scale experiments. By learning from historical battery designs, our approach can rapidly evaluate new battery designs without conducting time-consuming long-cycling experiments and even before cell prototyping. This achievement leads to substantial savings in both evaluation time and energy consumption compared to industrial practices and conventional data-driven workflows, thus enabling accelerated and energy-sustainable battery development. Beyond battery design, because DL is a generalizable ML workflow inherently designed to address the dual challenges of data unaffordability and distribution shift in AI4Science, we look forward to seeing further extension and application of DL to higher-level AI systems and other scientific domains, such as those for accelerating chemical research[53] and material synthesis[51].

**Discovery Learning**

*Inspiration and rationale*

To provide fast and reliable scientific prediction, an ideal ML workflow is expected to achieve high prediction accuracy while requiring minimal experimental costs for training and inference. Before this work, however, simultaneously addressing data unaffordability and distribution shift has remained an open question, which has hindered the realization of the promise of ML for accelerating scientific prediction and discovery. We hypothesized that an effective guiding principle—querying unlabeled test samples for labeling by effectively leveraging zero-cost historical battery data—can enable efficient scientific prediction by reducing both training and inference costs. In this study, the test samples refer to the previously unobserved combinations of electrode materials, cell designs, and cycling protocols. The historical battery data refers to existing public battery datasets of well-explored electrode materials and cell designs, as well as their degradation data with labeled cycle life. In this study, cycle life is defined as the number of charge and discharge cycles a battery can provide before its capacity drops to 90% of its initial value.

By exploring the human learning process and the factors behind human learning efficiency, we found that the learning theories in educational psychology have revealed the desired guiding principle. Specifically, we drew inspiration from Bruner's discovery learning theory proposed in the 1960s[52], which emphasizes an inquiry-based learning process supported by prior knowledge and past experiences. Further, this leads us to conclude that the efficiency of human reasoning could be attributed to the ability to generate new inferences based on prior inferences, rather than relying solely on direct observation.

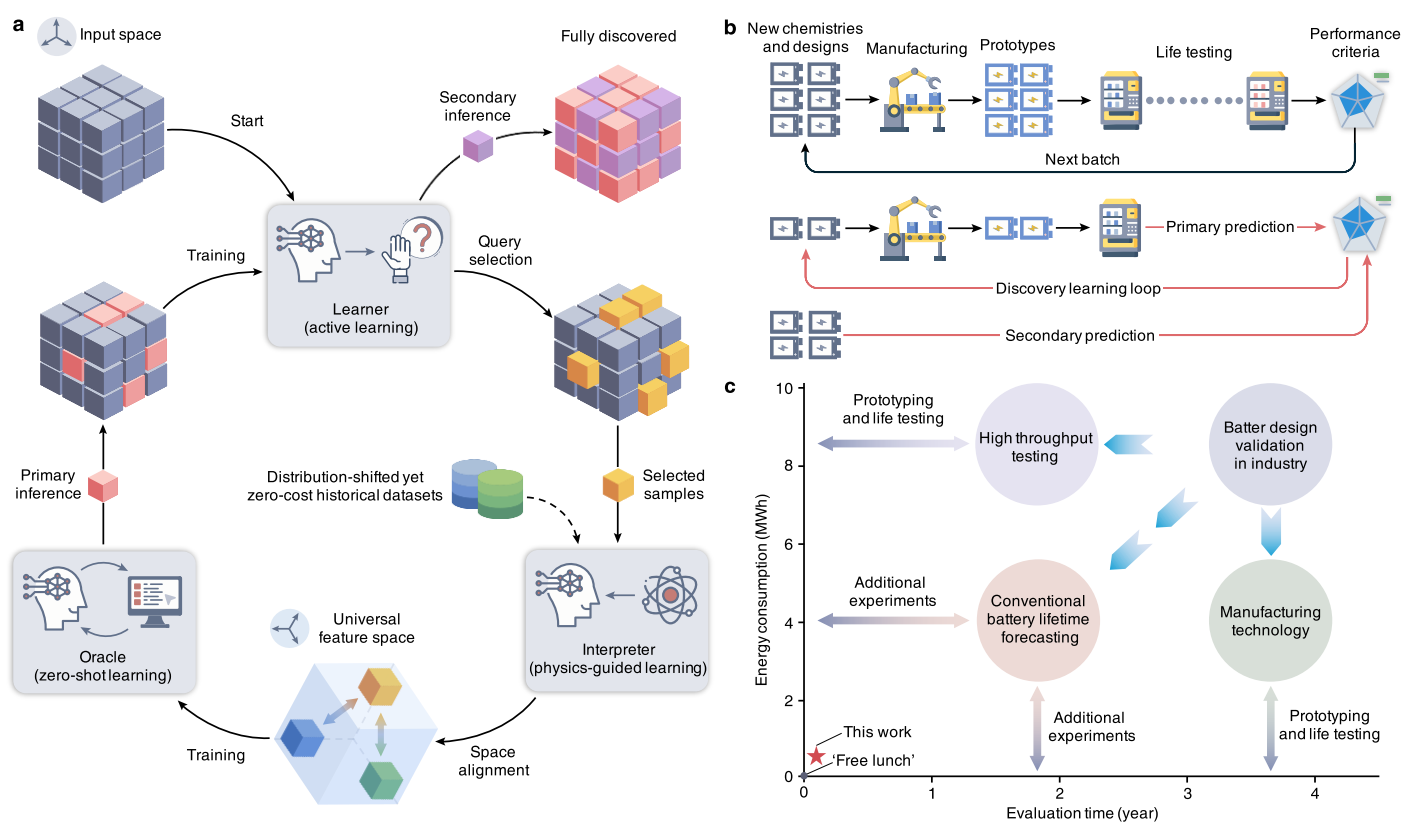

*Fig. 1 | Discovery Learning to accelerate battery design evaluation at an industrial scale. a*, Discovery Learning (DL) unites three core elements: the Interpreter, the Oracle, and the Learner. DL works as a human-like learning loop. Initially, the Learner actively makes queries by selecting more informative test samples. By 'test samples', we mean the previously unobserved combinations of cell chemistry, design, and cycling protocols. Next, the Interpreter constructs a universal physics-based feature space to align the new test samples with historical battery design datasets based on first-principles models. Following that, trained on historical datasets, the Oracle predicts test samples and provides the primary inference results to the Learner for training and the next-iteration query. Finally, the Learner performs the secondary inference. *b*, Application of DL in the context of industrial-scale battery design validation. The top figure shows the conventional industrial battery development process, including initial battery design, cell prototyping, and life testing, followed by comprehensively evaluating new battery designs based on all main performance criteria. The bottom figure illustrates how DL could be used in the practical battery industry to reduce evaluation time and energy consumption associated with battery prototyping and life testing. While the primary prediction can enable efficient early evaluation by learning from historical battery designs, the secondary prediction can achieve prototype-free evaluation by learning from primary predictions. While the costs of material mining and refining processes can be reduced as well, they are not included in the figures. The rapid battery lifespan evaluation can then enable fast feedback for battery design optimization. *c*, Figure of merit highlighting efficiency in terms of evaluation time and energy consumption. Under conservative assumptions (see Methods), the lifetime evaluation for 123 previously unobserved large-format pouch cells requires 1333 days (that is, approximately four years) and an energy consumption of 8.523 MWh. Under the same conditions, DL requires 33 days (that is, approximately four weeks) and an energy consumption of 0.468 MWh to finish the predictive evaluation of cycle-life performance with a mean-absolute-percentage error of 7.2% and a root-mean-squared error of 91 cycles. This achievement is equivalent to savings of 98% in evaluation time and 95% in energy consumption, demonstrating the effectiveness of DL. By eliminating upfront experimental costs for training and substantially reducing immediate experimental costs for inference, DL can outperform high-throughput testing, advanced manufacturing technologies, and conventional lifetime forecasting approaches.

## Concept of Discovery Learning

Here, we introduce DL, a flexible and scalable scientific ML approach that integrates active learning[5], physics-guided learning[6], and zero-shot learning[7] into an iterative and sequential reasoning process in a human-like manner (Fig. 1a). Corresponding to the three learning modules, we define DL's three agents: Learner, Interpreter, and Oracle. The DL process begins with the Learner actively selecting the most informative test samples without any labeled results (Fig. 1a). The Interpreter then leverages physics-guided learning to construct a universal and interpretable physics-based feature space, aiming to align feature discrepancies between historical batteries and unseen batteries. For example, these discrepancies can arise from differences in cell manufacturing. Next, the Oracle performs zero-shot learning, that is, learning solely from historical battery data without additional degradation experiments for cycle life testing. Specifically, it performs a primary inference on the selected test samples based on the feature space constructed by the Interpreter, and then provides the inference results as 'pseudo labels' to the Learner. Learning from the pseudo labels, the Learner uses active learning to select informative samples and assigns them to the Interpreter for the next round. A common strategy for evaluating the informativeness of a sample is to select the sample for which the model is least certain in its prediction[5]. This iterative and sequential reasoning loop continues until a predefined condition is met, for example, a number of iterations or a threshold of prediction uncertainty. At this point, the Learner finally performs a secondary inference on the remaining unselected samples.

For the Oracle and the primary inference, the training cost is zero, as the Oracle learns solely from historical battery data without requiring new degradation experiments for cycle life observations; however, the inference cost is necessary, as the Interpreter requires low-cost experiments (battery prototyping and early cycle-life testing) to construct the universal feature space. For the Learner and the secondary inference, the training cost is exactly Oracle's inference cost and can be further reduced by the Learner through active learning, as the Learner learns solely from Oracle's primary inference results; the inference cost is zero, as the secondary inference is based on independent features predetermined before experiments (that is, cycling protocols). Therefore, DL is inherently designed to fully eliminate upfront training costs by effectively leveraging zero-cost historical datasets and to substantially reduce immediate inference costs by actively selecting more informative test samples.

*Rapid and sustainable battery design evaluation*

In this work, DL is applied to accurately evaluate the cycle life of previously unobserved battery designs even under unknown device variability, without requiring new degradation experiments for labeling lifetime and with substantially reduced prototyping for extracting early features. The objective is to enable efficient predictive modeling of battery degradation by minimizing the time and energy costs required for prototyping and life testing (Fig. 1b). Fig. 1c presents a figure of merit highlighting experimental costs in terms of time and energy. In this study, we demonstrate that DL can achieve a mean-absolute-percentage error (MAPE) of 7.2% in predicting the cycle life of previously unobserved battery designs with unknown device variability, using only the first 50 equivalent full cycles (EFCs) from just 51% of the cell prototypes (see Fig. S9 and Fig. S11). It is important to emphasize that this high accuracy is achieved in a zero-shot manner, which even surpasses representative achievements on well-explored battery designs[3] and significantly outperforms state-of-the-art results of few-shot lifetime prediction[31]. Under conservative assumptions, DL achieves savings of 98% in evaluation time and 95% in energy consumption compared to industrial battery lifespan validation, that is, from nearly four years and 8.5 MWh to four weeks and 0.5 MWh (see Methods). DL achieves a rapid validation of battery lifespan and enables fast and high-throughput feedback for battery design, which offers a promising step toward accelerating battery innovation and addressing the 'sustainability dilemma'.

*Implementation of Discover Learning*

*Physics-guided learning*

The task of the Interpreter is to construct a universal and physically interpretable feature space through physics-guided learning. We applied simulation-based inference techniques[54,55] to sample the probability distributions of 14 physical parameters based on a physics-based electrochemical battery model and early cycling profiles (see Fig. S5-S6). These parameters are related to thermodynamic and kinetic properties during early cycling. We can then obtain 28 physics-based statistical features: 14 mean values from the first cycle and 14 incremental values over the first 50 EFCs (see Fig. 3c caption).

*Zero-shot learning*

With the features derived by the Interpreter, the task of the Oracle is to perform primary inference through zero-shot learning. We hypothesized that the key to successfully achieving zero-shot prediction is to learn generalizable knowledge from historical battery data. Inspired by the idea of meta-learning[56,57], we developed a dual-predictor architecture consisting of a base predictor and a meta-predictor, decoupling the effects of cycling conditions and physics-based features (see Fig. S7). While the base predictor directly establishes the relationship between physics-based features and cycle life, the meta-predictor learns a 'meta-level knowledge'—that is, how cycling conditions influence the importance of initial physical states and early-cycle physical evolutions (see Supplementary Information for more details).

*Active learning*

Based on the primary inference from the Oracle, the task of the Learner is to perform active learning, including query and secondary inference. We combined two types of query strategies to select informative test samples: (i) unsupervised rule-based query strategy, which is based solely on cycling-condition features without the need for pseudo labels from the Oracle; it is used at the beginning of the DL process; and (ii) supervised uncertainty-based query strategy, which requires pseudo labels from the Oracle to determine which unlabeled samples have higher prediction uncertainty (see Fig. S12). In this study, we applied a Gaussian process regression model[58] to evaluate prediction uncertainty and perform secondary inference on cycle life using only cycling-condition features.

**Results**

*Industrial-grade large-format battery aging data*

To test the effectiveness of DL and to contribute to battery research, we present an industrial-grade battery degradation dataset comprising 123 large-format (73–84 Ah) lithium-ion pouch cells. This dataset spans eight distinct cell types with varied electrode materials and cell designs, including next-generation silicon-based anode and ultrahigh-nickel cathode materials (see Table S1). To comprehensively verify DL, cycle life testing was conducted under diverse cycling conditions (see Fig. S1-S4). To our knowledge, no existing public dataset provides comprehensively validated data on large-format batteries (>10 Ah[59]). Despite the critical role of large-format batteries in enhancing battery integration efficiency and increasing EV cruise range[60], their degradation behaviors remain insufficiently studied due to data scarcity. Hence, our dataset can contribute to the battery community and facilitate further exploration of large-format batteries. Further, we constructed the training set by collecting zero-cost public datasets consisting of small-capacity (1.1–3.5 Ah) cylindrical cells, which differ from our test dataset in electrode material, cell design, and manufacturing (see Table S1 for more details).

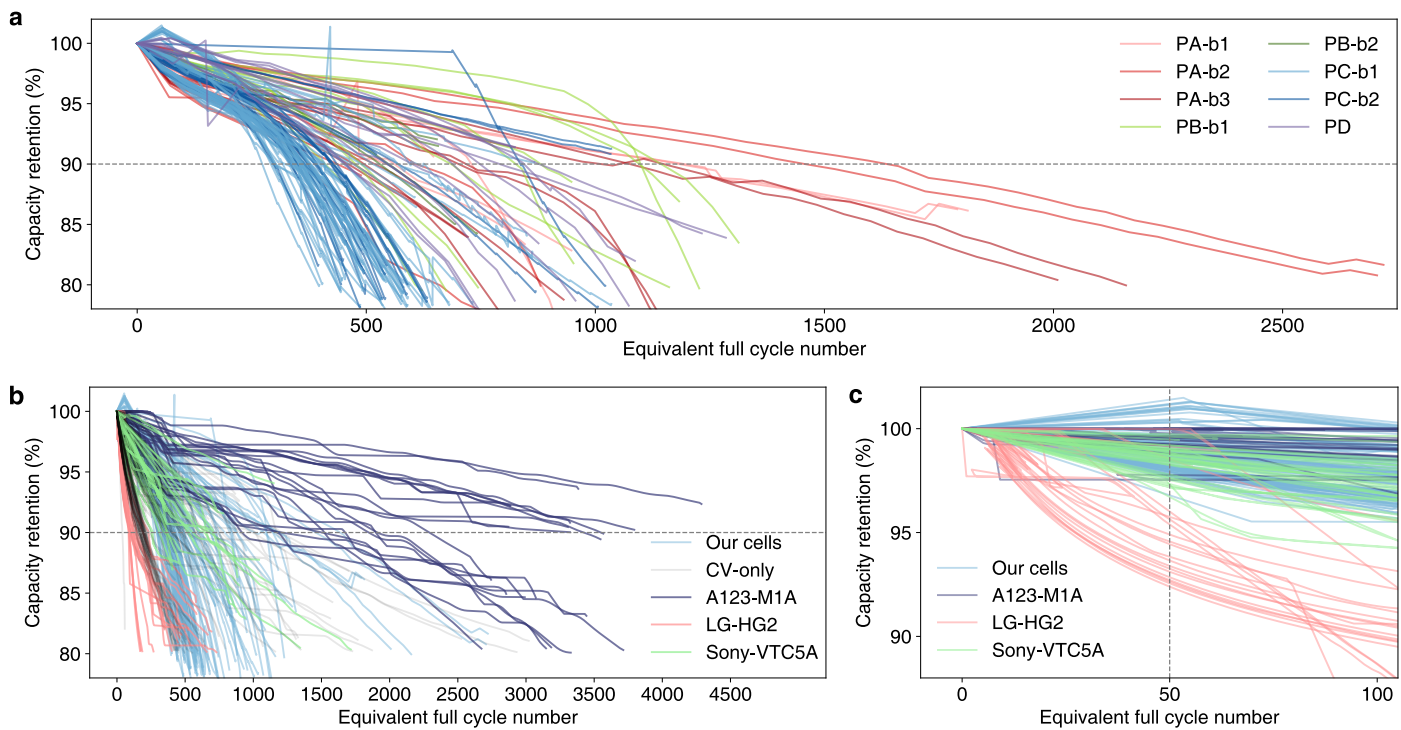

*Fig. 2 | Overview of industrial-grade battery aging data of large-format pouch cells and other public battery datasets of small-capacity cylindrical cells. **a**, The retention of C/3 discharge capacity versus equivalent full cycle number for our large-format lithium-ion pouch cells, including eight types spanning different material-design combinations. In line with battery industry practice, minor variations in materials and cell designs resulting from different production batches are treated as distinct cell types. For example, PA-b1 and PA-b2 are different batches. 'P' denotes pouch cell, and 'A' is an arbitrary code name. **b**, Overview of all battery aging data used in this study. We collected public datasets consisting of small-capacity cylindrical lithium-ion cells from six commercial models: A123-M1A, LG-HG2, Sony-VTC5A, LG-MJ1, Sony-VTC6, and Samung-25R. Among them, A123-M1A, LG-HG2, and Sony-VTC5A are used throughout the study, while the other three models are used only for cross-validation (CV) during hyperparameter selection and excluded from training by the Oracle due to data-filtering conditions (see Supplementary Information 'zero-shot learning'). **c**, A detailed view of **b** without CV-only cells, showing the first 50 equivalent full cycles used in this study.*

Fig. 2a provides an overview of our test dataset, with cycle lives ranging from 250 to 1700 cycles. Fig. 2b clearly shows that our dataset and public datasets exhibit different degradation behaviors. Public batteries mostly show a linear or gradually decelerating decline in capacity retention, whereas most of our cells exhibit a transition from linear to accelerated degradation. This deviation implies a difference in dominant degradation mechanisms. Furthermore, Fig. 2c offers a close-up of the first 50 EFCs, highlighting another key difference. While public cells exhibit either an early capacity drop or a relatively flat capacity trend, some of our cells exhibit an initial capacity increase. A possible reason for this behavior is the mechanical constraint of fixed-volume fixtures with foam padding during cycling (see Fig. S13). Under such conditions, early electrode expansion may enhance contact between solid materials, thereby reducing internal resistance and increasing discharge capacity. These differences in degradation behavior highlight the challenge of predicting the lifespan of unseen battery designs.

To consider device variability due to manufacturing, we conducted repeated life testing for identical cell types under the same cycling conditions. Accordingly, the 123 cells were divided into 37 groups, each comprises cells with highly consistent electrode chemistry, cell design, and cycling conditions (see Fig.S8-S9). Based on this grouping, our ultimate objective is to predict the average cycle life within each of the 37 cell groups, with final prediction accuracy evaluated at the group level.

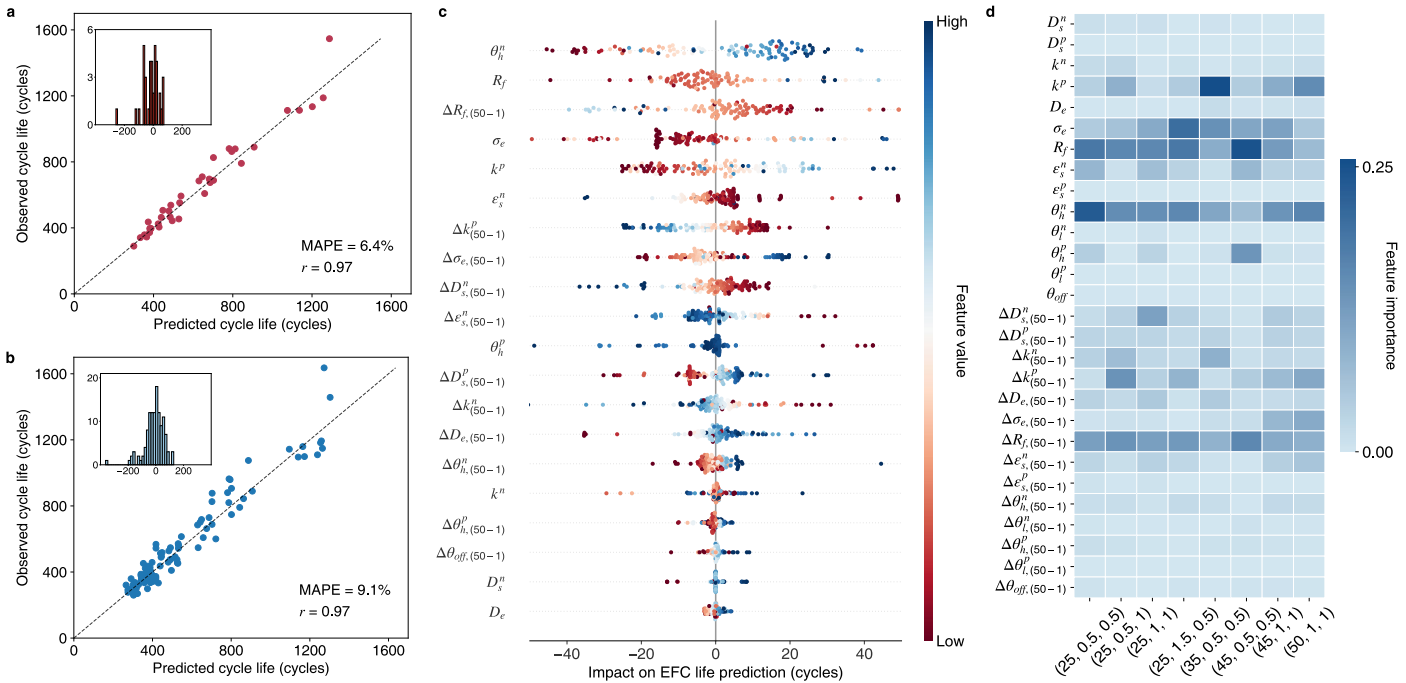

***Fig. 3 | Results of open-loop DL from the Interpreter and the Oracle without the Learner in the loop. a,*** *The average cycle life within each of the 37 cell groups: observed results from experimental validation versus predicted results from the Oracle, assisted by the Interpreter. The MAPE on 37 cell groups is 6.4%, and the Pearson correlation coefficient r is 0.97.* ***b,*** *The cycle life of the 123 individual cells: observed results from experimental validation versus predicted results from the Oracle, assisted by the Interpreter. The MAPE on 123 individual cells is 9.1%, and the Pearson correlation coefficient r is 0.97.* ***c,*** *Impact of key physics-based features on predicting the cycle life. Each data point corresponds to a cell. There are 28 physics-based features in total, where 14 of them are the means of the parameter distribution results at the first cycle—representing the initial physical states—and the other 14 of them are the variations in these means over the first 50 equivalent full cycles—representing the evolution of physical states. The nomenclature for the 14 physics-based features at the first cycle: (1) '$D_s^n$' refers to the diffusivity in solid active materials at negative electrode; (2) '$D_s^p$' refers to the diffusivity in solid active materials at positive electrode; (3) '$k^n$' refers to the lithium intercalation or deintercalation reaction rate constant at negative electrode; (4) '$k^p$' refers to the lithium intercalation or deintercalation reaction rate constant at positive electrode; (5) '$D_e$' refers to the diffusivity of the electrolyte; (6) '$\sigma_e$' refers to the ionic conductivity of the electrolyte; (7) '$R_f$' refers to the lumped film resistance of the solid electrolyte interphase at negative electrode; (8) '$\varepsilon_s^n$' refers to the active-material volume fraction at negative electrode; (9) '$\varepsilon_s^p$' refers to the active-material volume fraction at positive electrode; (10) '$\theta_h^n$' refers to the lithium stoichiometry at the upper cutoff cell voltage (that is, higher voltage limit) at negative electrode; (11) '$\theta_l^n$' refers to the lithium stoichiometry at the lower cutoff cell voltage at negative electrode; (12) '$\theta_h^p$' refers to the lithium stoichiometry at the upper cutoff cell voltage at positive electrode; (13) '$\theta_l^p$' refers to the lithium stoichiometry at the lower cutoff cell voltage at positive electrode; (14) '$\theta_{off}$' refers to the offset in lithium stoichiometry at the lower cutoff cell voltage between negative electrode and positive electrode. Further, '$\Delta_{(50-1)}$' refers to the difference in the mean value between the fiftieth equivalent full cycle and the first cycle. We note that features with negligible impact have been excluded from the figure.* ***d,*** *Variation in the importance of physics-based features for predicting cycle life under different cycling conditions. The cycling-condition features are described by the '(ambient-temperature feature, charge C rate feature, discharge C rate feature)'. For example, the '(25, 0.5, 0.5)' represents that the ambient-temperature feature is 25 °C, the charge C rate feature is 0.5 C, and the discharge C rate feature is 0.5 C. We note that cycling conditions with insufficient sample sizes for evaluating feature importance have been excluded from the figure.*

## *Open-loop results from the Oracle*

We first evaluate the open-loop predictive performance of DL, and the results shown in Fig. 3a demonstrate the effectiveness of the primary inference. The Oracle achieved a 6.4% group-level MAPE in predicting the average cycle life within each of the 37 cell groups under unknown manufacturing variability, while the root-mean-squared error (RMSE) is 64 cycles. Furthermore, a 9.1% cell-level MAPE (RMSE is 70 cycles) (Fig. 3b) illustrates that the 6.4% group-level MAPE is largely driven by accurate predictions of the 123 individual cells, rather than being a coincidence resulting from within-group averaging. Additionally, the Pearson correlation coefficient of 0.97 further verify the predictive capability of our approach. We then use SHapley Additive exPlanations analysis to illustrate the relative importance of physics-based features for predicting cycle life (Fig. 3c). For example, it can be seen that a lower 'active-material volume fraction at negative electrode' $\varepsilon_s^n$ or a larger 'lithium stoichiometry at the upper cutoff cell voltage at negative electrode' $\theta_h^n$ tends to have a positive impact on battery cycle life. The larger $\theta_h^n$ generally corresponds to a broader stoichiometry range, which can be equivalent to a lower $\varepsilon_s^n$ for a given electrode capacity. Meanwhile, the $\varepsilon_s^n$ is proportional to the Brunauer–Emmett–Teller area (that is, specific surface area) of active materials. Therefore, both lower $\varepsilon_s^n$ and larger $\theta_h^n$ represent a lower Brunauer–Emmett–Teller area, which is a well-known factor and a common cell design strategy for extending battery lifespan[60]. Further, Fig. 3d shows the variation in the importance of physics-based features under different cycling conditions. For example, the results show that the 'lumped film resistance at negative electrode' $R_f$ has a higher importance under 45 °C and 0.5 C. This is consistent with the properties of solid-electrolyte-interphase growth side reaction, which dominates the loss of lithium inventory at elevated temperatures.

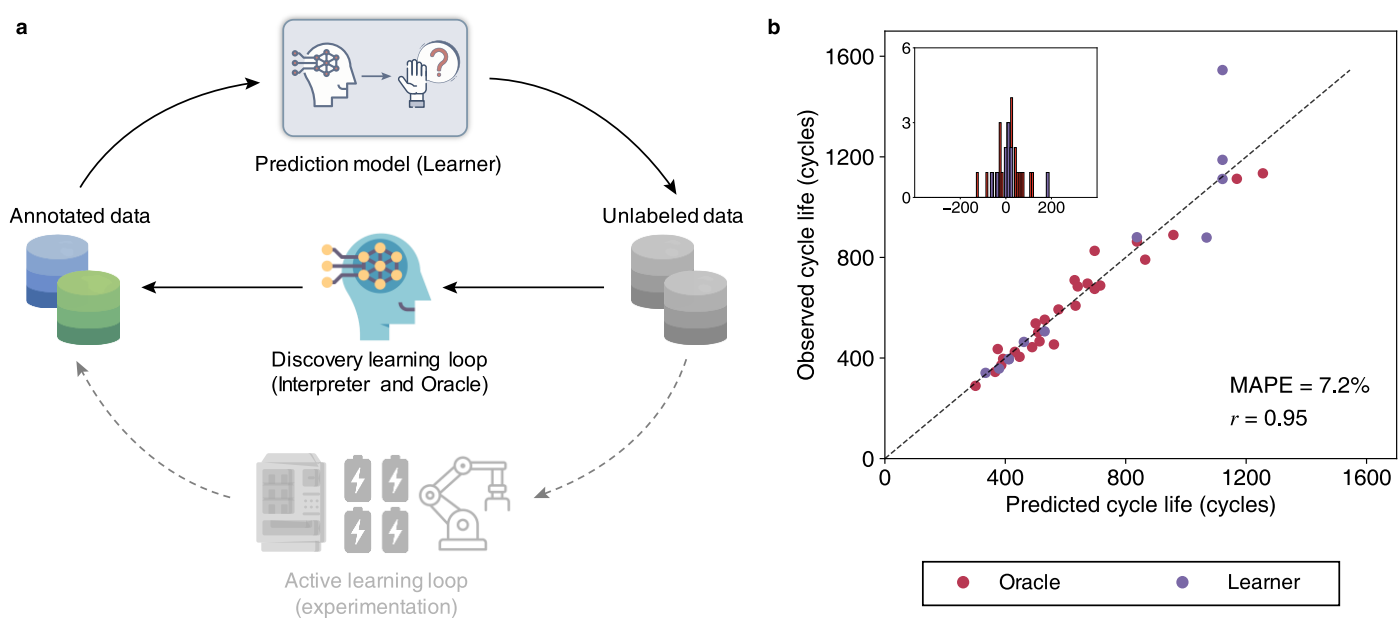

***Fig. 4 | Results of closed-loop DL with the Learner in the loop. a,** The DL loop versus conventional active-learning loop. For a conventional pipeline of active query and learning, the data-labeling (or data annotation) process relies on experimentation. In contrast, the DL loop uses the prediction results of the Oracle (that is, the primary inference results) based on zero-shot learning to replace the experiment-driven data labeling, thus eliminating the training costs of the Learner. **b,** The average cycle life within each of the 37 cell groups: observed results from experimental validation versus predicted results from both the Oracle and the Learner, assisted by the Interpreter. The MAPE on 37 cell groups is 7.2% (Pearson correlation coefficient r=0.95), which is slightly higher than the open-loop results (6.4%) predicted by only the Oracle. This is a normal phenomenon because the collaboration of the Oracle and the Learner will inherently cause error accumulation.*

## *Closed-loop results from both the Oracle and the Learner*

While the Oracle can perform accurate zero-shot prediction, the experimental costs required by the Interpreter may cause critical bottlenecks in large-scale predictions. To address this issue, we incorporate the active-learning strategy into the Learner, to actively select more informative test samples from the 37 cell groups (each group is a sample). Fig. 4a clearly illustrates the difference between the DL loop and the active-learning loop: DL replaces the experimental labels in active learning with the pseudo labels provided by the Oracle through primary inference. To ensure more stable predictive performance given the small sample size of 37 cell groups, we allocated a larger portion of the query budget to the unsupervised strategy (22 cell groups), applying a 70% upper limit within each cell type (except for the cell types that contain only two groups; see Fig. S10-S11). For example, the unsupervised query will select three cell groups from the cell type that includes five groups. Following the unsupervised query, we conducted one round of supervised querying. As a result, the Learner identified four informative samples with a prediction standard deviation larger than the upper quartile (see Fig. S12). As such, the Learner selected a total of 26 cell groups. With the help of the Oracle, which achieves a 7.1% group-level MAPE (RMSE is 56 cycles) for predicting the selected 26 groups, the Learner achieves a 7.4% group-level MAPE (RMSE is 142 cycles) for predicting the remaining 11 groups. Fig. 4b shows that the Oracle and the Learner together achieve a 7.2% group-level MAPE (RMSE is 91 cycles) in predicting the average cycle life within each of the 37 cell groups, demonstrating the closed-loop effectiveness of DL.

**Discussion**

We successfully demonstrated the effectiveness of DL in performing rapid predictive validation of battery lifespan. DL can also be integrated with advanced manufacturing technologies[24] and high-throughput testing to reduce time and energy costs further. Furthermore, provided that suitable historical datasets exist, DL may be extended to other battery performance criteria, such as safety[61] and fast charging capability[62]. Looking ahead[12], as data volumes expand, DL could unlock the potential of the Battery Data Genome project by extracting valuable information from historical battery datasets. Beyond battery design, successful deployment of battery technologies also requires advancements in other sectors, such as battery management[63] and second-life repurposing[64]. How to extend DL to these areas deserves further investigation. This study highlights the promise of uncovering insights from historical battery design concepts to inform and accelerate the development of next-generation technologies and solutions.

More broadly, DL offers a novel concept in AI4Science and represents a paradigm shift toward more efficient data-driven modeling (Fig. 5). As a scalable and generalizable framework, we anticipate that DL will be a foundation for more advanced scientific prediction models. For example, a state-estimation technique for physical systems without known governing equations[65] can be integrated into DL to enhance generalization ability. Furthermore, by leveraging advancements in experimental characterization, such as non-destructive internal temperature measurements[66] and pixel-by-pixel image data[38], DL offers new possibilities for the efficient exploration of complex physical systems where data is not readily accessible. Moreover, DL has the potential to overcome the limitations of existing AI-guided closed-loop optimization strategies by substantially reducing experimental costs associated with early predictors[4] or closed-loop experiments[51,53,67,68], which can help realize the promise of ML for accelerating scientific discovery and engineering innovation (see Fig. 5 and Fig. S14-S15).

Beyond AI4Science, there is increasing attention on artificial general intelligence[69], where a central question is how machine intelligence can achieve human-like learning and reasoning[56,70]. The concept of DL represents an example of mimicking human learning processes, highlighting the promise of incorporating human learning theories into the design of ML methods to advance human-level AI. Additionally, the efficiency-related motivation behind DL aligns with one of the core goals of artificial general intelligence: developing ML systems that can learn as efficiently as humans and animals. Hence, we look forward to seeing DL's potential fully realized in this direction.

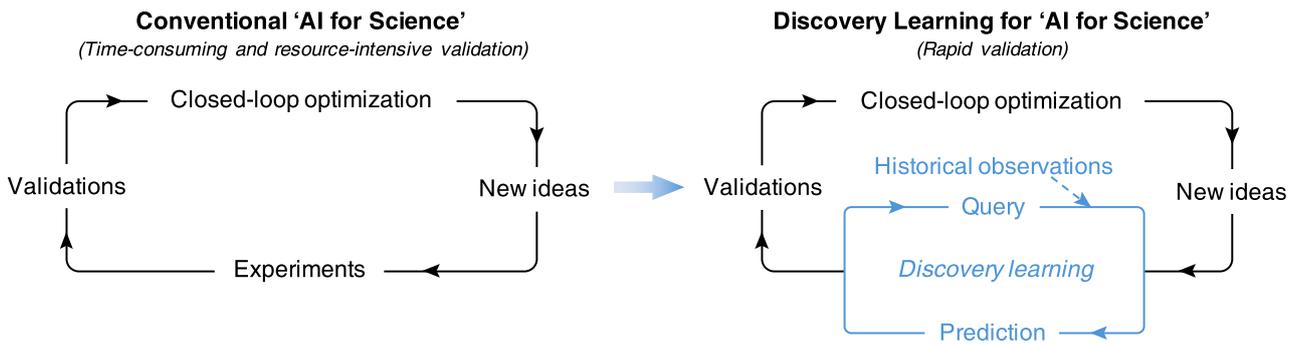

**Fig. 5 | DL enhances AI4Science.** AI-guided closed-loop optimization has been widely applied in scientific domains, and the overall exploration efficiency is bottlenecked by the forward validation (evaluation) process. By learning from zero-cost historical observations and actively reducing inference costs, DL offers a promising predictive approach to accelerating the general validation process of new scientific hypotheses and technologies. This represents a paradigm shift in scientific discovery and engineering innovation toward higher efficiency. By replacing time-consuming and resource-intensive experimental validation with rapid predictive evaluation, DL helps realize the promise of AI and ML for empowering a broad range of scientific and engineering domains that involve complex dynamical systems, costly experimentation, and large-scale design space.

## Methods

### Cell chemistry and design

The test dataset comprises 123 large-format lithium-ion pouch cells (73–84 Ah) with two types of cathode materials (positive electrode): NMC811 and blended NMC9 (a 1:1 weight ratio blend of NMC9 and NMC811). Two anode chemistries (negative electrodes) are also used: graphite and Si-C composite (containing 10 wt% silicon). In addition to electrode chemistries, the inactive electrode materials (that is, conductive additions and binders) and electrolyte chemistries can be different; while further details are not disclosed due to commercial confidentiality, they are not central to the objectives or conclusions of this study.

Moreover, variations in cell design include parameters such as electrode thickness, porosity, and particle size. Due to the proprietary nature of the data, the exact values of these key design parameters cannot be made publicly available; instead, the relative values in comparison with the cell design parameters of public datasets will be disclosed (see Supplementary Information for more details). Furthermore, consistent with industry practice, minor design variations due to cost control—such as inter-batch differences in manufacturing—are also treated as distinct cell types. For example, PA-b1, PA-b2, and PA-b3 share the same chemistry and basic cell design, but originate from different manufacturing batches. The minor batch-to-batch differences are not disclosed due to commercial confidentiality and are not essential to the core findings of this study.

### Cycle life testing

To comprehensively evaluate cycle life performance, the cycle life testing involves three types of cycling protocols: (1) the standard, ell-accepted CC discharge profile; (2) the dynamic discharge profile representative of EV driving; and (3) the multi-step charge profile comprising several CC steps followed by a CV step. Cycle life tests were performed on large-format pouch cells under diverse operating conditions, with variations in ambient temperature and in both charge and discharge C rates. In addition, the reference performance tests (RPTs, see Fig. S4 in Supplementary Information), involving C/3 CC discharge testing protocols, were periodically inserted throughout the cycling process. (see Supplementary Information for more details)

### Public battery datasets

We collected public datasets consisting of 200 small cylindrical cells (1.1–3.5 Ah) from six commercial models: A123-M1A, LG-HG2, Sony-VTC5A, LG-MJ1, Sony-VTC6, and Samung-25R. These datasets were published as part of previous studies or projects related to battery degradation[71–75]. In addition, we selected these public datasets based on the availability of cell design information, which was searched and obtained from the previous literature[76,77]. Among them, A123-M1A, LG-HG2, and Sony-VTC5A are used throughout the study, while the remaining three models are used only for cross-validation during hyperparameter selection and are excluded from the training of the Oracle based on predefined usage criteria.

### Physics-based electrochemical model

Open research software, PyBaMM[78], was used to build and solve the physics-based electrochemical model. The governing equations and details are shown in the Supplementary Information equations (1)–(8). The modeling parameters used in this study can be found in the relevant code, which will be made available upon final publication. Conceptually, the material properties associated with cell electrode chemistries (such as the open-circuit potential characteristics) and basic cell design parameters (such as thickness, particle radius, and porosity) are predetermined and fixed, while the remaining parameters related to thermodynamics and kinetics are set as unknown variables, the posterior probability distributions of which are the targets of the Interpreter based on physics-guided learning. (see Supplementary Information for more details)

*Physics-guided learning*

The Interpreter in DL is based on physics-guided learning; specifically, the simulation-based inference technique was applied in this study (see Fig. S5 in Supplementary Information). Open Python package, 'sbi'[79], was used to implement the simulation-based inference technique. Combined with the physics-based electrochemical model, the first step is to use 'sbi' to randomly generate data samples from the physics-based parameter space (comprising 11 physical parameters). To balance computational efficiency and data robustness, we set the sample size to 50000. Next, these generated data were used to train a neural network-based conditional density estimator[80], which was developed by applying the neural spline flow model[81]. The neural spline flow model takes CC charge or discharge profiles as input and produces parameter posterior distributions as output. Finally, the trained model was used to generate the posterior distribution of the 11 parameters (see Fig. S6); subsequently, these 11 parameters were extended to 14 parameters with three additionally derived, dependent parameters (see Supplementary Information for more details). We note that cell types with the same electrode chemistries and basic design parameters (including thickness, particle radius, and porosity, while excluding design information such as electrolyte chemistries) share the same neural spline flow model.

*Zero-shot learning*

The Oracle in DL is designed to perform zero-shot learning; specifically, a dual-predictor architecture comprising a base predictor and a meta-predictor was developed in this study (see Fig. S7 in Supplementary Information). The base predictor takes physics-based features as input and produces battery cycle life as output; a linear model combined with the elastic net[82] was applied to develop the base predictor. The meta-predictor takes cycling conditions as input and produces the weight coefficients of physics-based features as output; a support vector regression model[83] was applied to develop the meta-predictor. Open Python package, Scikit-learn[84], was used to implement the linear model and the support vector regression model. (see Supplementary Information for more details)

*Active learning*

The Learner in DL is designed to perform active query and learning. Specifically, the prediction model of the Learner was developed by applying a Gaussian process regression model[58], and Scikit-learn was used to implement this prediction model. This prediction model takes cycling conditions as input and produces battery cycle life as output. Two types of query strategies were applied, including the unsupervised strategy and the supervised query strategy.

The unsupervised query strategy is based on the predefined rules within each cell group associated with cycling-conditions features (that is, ambient temperature, charge C rate, and discharge C rate): under the query budget limit, (1) the first criterion is to maximize the diversity in the ambient-temperature dimension while ensuring the boundary samples are selected with higher priority; (2) when a choice needs to be made between samples with the same ambient-temperature features, the second criterion is to select the samples with lower C rate (a higher priority is given to the charge C rate) for higher ambient temperature, or the samples with higher C rate for lower ambient temperature, based on the prior knowledge about battery degradation paths; and (3) the third criterion is to maximize the diversity in the C rate dimensions (a higher priority is given to the charge C rate) while ensuring the boundary samples are selected with higher priority.

Furthermore, the supervised query strategy is based on the prediction uncertainty of samples, which is calculated by the Gaussian process regression model. Specifically, test samples with prediction standard deviation exceeding the upper quartile are selected for primary inference by the Oracle, while the remaining samples undergo secondary inference by the Learner. (see Supplementary Information for more details)

*Machine-learning model evaluation*

MAPE and RMSE are chosen to evaluate the predictive performance of ML models.

MAPE (in units of %) is defined as

$$\text{MAPE} = \frac{1}{n}\sum_{i=1}^{n}\frac{|\hat{y}_i - y_i|}{y_i} \times 100$$

where $\hat{y}_i$ is the predicted cycle life, $y_i$ is the observed cycle life, and $n$ is the total number of test samples.

RMSE (in units of cycles) is defined as

$$\text{RMSE} = \sqrt{\frac{1}{n}\sum_{i=1}^{n}(\hat{y}_i - y_i)^2}$$

where all variables are defined as above.

*Experimental costs in time and energy*

We evaluate the experimental costs of DL in terms of time span and energy consumption compared to industrial battery design validation. For the calculation of the time costs, we have two sets of assumptions. The first set of assumptions include: (1) assuming an average equivalent full cycle life of 1000 cycles; (2) assuming a 1 C charge and discharge, which is equivalent to two hours for one charge-discharge cycle; (3) assuming that the battery tester can simultaneously test eight cells in one round. Under assumptions (1) and (2), the full life testing requires 2000 hours, that is, 83.3 days; the first-50-cycles testing requires 100 hours, that is, 4.2 days. Furthermore, under assumption (3), the 123 cells require 16 rounds of the full life testing, that is, 1333.3 days; the 63 cells selected by DL require eight rounds of the first-50-cycles testing, that is, 33.3 days.

The second set of assumptions include: (1) assuming an average equivalent full cycle life of 1000 cycles; (2) assuming a 2 C charge and discharge, which is equivalent to one hour for one charge-discharge cycle; (3) assuming that the battery tester can simultaneously test four cell groups in one round. Under assumptions (1) and (2), the full life testing requires 1000 hours, that is, 42 days; the first-50-cycles testing requires 50 hours, that is, two days. Furthermore, under assumption (3), the 37 cell groups require ten rounds of the full life testing, that is, 420 days; the 26 cell groups selected by DL require seven rounds of the first-50-cycles testing, that is, 14 days.

For the calculation of the energy costs, several assumptions were made: (4) conservatively assuming the energy consumption per unit in prototype manufacturing as the average of the lower bound of energy consumption for lithium-ion battery (that is, 20.3 kWh$_{prod}$ per kWh$_{cell}$) and post lithium-ion battery (that is, 10.6 kWh$_{prod}$ per kWh$_{cell}$) production[2], that is, 15.45 kWh$_{prod}$ per kWh$_{cell}$ after averaging; (5) assuming an average cell energy capacity of 275 Wh (the capacity is calculated by averaging the lower bound 73 Ah and the upper bound 84 Ah; the voltage is assumed as 3.5 V); and (6) assuming that the battery tester demands power of 250 W[85]. Under assumptions (4) and (5), the energy consumption for manufacturing the 123 large-format cells is 523 kWh; the energy cost of DL for manufacturing the selected 63 cells is 268 kWh.

Under the first set of assumptions (1), (2), (3), and (6), the energy consumption in the 16 rounds of the full-life testing is 8000 kWh; the energy cost of DL in the eight rounds of the first-50-cycles testing is 200 kWh. In summary, the energy consumption of industrial battery design validation is 8.523 MWh, while the energy consumption of DL is 0.468 MWh. Under the second set of assumptions (1), (2), (3), and (6), the energy consumption in the ten rounds of the full-life testing is 2500 kWh; the energy cost of DL in the seven rounds of the first-50-cycles testing is 88 kWh. In summary, the energy consumption of industrial battery design validation is 3.023 MWh, while the energy consumption of DL is 0.356 MWh.

*Reproducibility.*

All results and figures in this study can be reproduced using the provided code or by following the method descriptions.

**Data availability**

The data originally presented in this study will be made publicly available upon final publication. The raw data of cycle life testing will be made available. Due to commercial confidentiality, certain commercially sensitive cell design data, which are not central to the conclusions of this study, will be made available in a processed form. The public datasets used in this study are accessible online or available upon request from the original authors. The A123-M1A dataset and LG-HG2 dataset are from Sadia National Laboratories[71]. The Sony-VTC5A dataset is from Technical University of Munich[74]. The LG-MJ1 dataset is from Flemish Institute for Technological Research[72]. The Sony-VTC6 dataset is from Carnegie Mellon University[75]. The Samung-25R dataset is from Karlsruhe Institute of Technology[73].

**Code availability**

The code developed in this study will be made publicly available upon final publication.

**Acknowledgements**

This study was supported by Farasis Energy USA. The large-format lithium-ion pouch cells and all relevant data were provided by Farasis Energy USA. We thank Q. Hu for early-stage data processing. We thank B. Li and W. Xu for early-stage processing and analysis of public datasets. We also thank all referenced articles and their authors for providing the motivation, justification, and inspiration for this study.

**Author contributions**

J.Z. and Z.S. conceived the study and led the research program. J.Z. developed the Discovery Learning concepts, including the physics-guided learning approach, the zero-shot learning approach, and the active learning approach. W.J., Y.R., and Q.J. provided experimental data and performed data management. J.Z., Y.R., and Q.J. interpreted the experimental data. J.Z. and Y.Z. developed the algorithms for physics-guided learning and zero-shot learning. J.Z. and B.Y. developed the algorithm for active learning. J.Z. and H.B. performed experimental cost analysis. Y.Z. and J.Z. performed code management. J.Z. reviewed previous literature and wrote the paper. J.Z., Y.Z., and B.Y. prepared the visualization items. J.Z., Y.Z., B.Y., and H.B.

prepared the supplementary material. All authors edited and reviewed the manuscript. W.J. and Z.S. supervised the work.

## Competing interests

The authors declare no competing interests.

## Additional information

*Supplementary information.* The online version containing supplementary material will be made available upon final publication.

*Correspondence and requests for materials* should be addressed to W.J. or Z.S.